\documentclass[11pt, a4paper]{article}

\usepackage[margin=1in]{geometry}
\usepackage{amsmath, amssymb, amsthm}
\usepackage{graphicx}
\usepackage{algorithm}
\usepackage{algorithmic}
\usepackage{hyperref}
\usepackage{cite}
\usepackage{color}
\usepackage{soul}
\usepackage{enumitem}
\usepackage{booktabs}
\usepackage{multirow}
\usepackage{tikz}
\usepackage{pgfplots}
\pgfplotsset{compat=1.17}
\usetikzlibrary{shapes.geometric, arrows}
\usepackage{fancyhdr}

\theoremstyle{definition}

\title{\textbf{LLM Powered Social Digital Twins: \\ A Framework for Simulating Population Behavioral Response to Policy Interventions}}

\author{
  \centering
  \begin{minipage}[t]{0.3\textwidth}
    \centering
    \textbf{Dr Fatima Koaik}\
    \texttt{fatima.koaik@pwc.com}
  \end{minipage}
  \hfill
  \begin{minipage}[t]{0.3\textwidth}
    \centering
    \textbf{Aayush Gupta}\
    \texttt{aayush.brij.gupta@pwc.com}
  \end{minipage}
  \hfill
  \begin{minipage}[t]{0.5\textwidth}
    \centering
    \textbf{Farahan Raza Sheikh}\
    \texttt{farahan.raza.sheikh@pwc.com}
  \end{minipage}
}
\date{January 3, 2026}

\begin{document}

\maketitle

\begin{abstract}
Predicting how populations respond to policy interventions is a fundamental challenge in computational social science and public policy. Traditional approaches rely on aggregate statistical models that capture historical correlations but lack mechanistic interpretability and struggle with novel policy scenarios. We present a general framework for constructing \textbf{Social Digital Twins}  virtual population replicas where Large Language Models (LLMs) serve as cognitive engines for individual agents. Each agent, characterized by demographic and psychographic attributes, receives policy signals and outputs multi dimensional behavioral probability vectors. A calibration layer maps aggregated agent responses to observable population level metrics, enabling validation against real world data and deployment for counterfactual policy analysis.

We instantiate this framework in the domain of pandemic response, using COVID-19 as a case study with rich observational data. On a held out test period, our calibrated digital twin achieves a \textbf{20.7\% improvement} in macro averaged prediction error over gradient boosting baselines across six behavioral categories. Counterfactual experiments demonstrate monotonic and bounded responses to policy variations, establishing behavioral plausibility. The framework is domain agnostic: the same architecture applies to transportation policy, economic interventions, environmental regulations, or any setting where policy affects population behavior. We discuss implications for policy simulation, limitations of the approach, and directions for extending LLM based digital twins beyond pandemic response.
\end{abstract}

\section{Introduction}
\pagestyle{plain}

Governments continuously face the challenge of predicting how populations will respond to proposed policies. Will a carbon tax reduce driving, or will people simply absorb the cost? How will citizens react to a new social welfare program? Will businesses comply with environmental regulations? These questions are fundamentally about \textit{behavioral simulation}  modeling how heterogeneous individuals make decisions in response to changing incentives and constraints.

Traditional approaches to policy prediction fall into two categories. \textbf{Aggregate statistical models} (econometric regressions, time series forecasting) capture historical correlations but lack mechanistic interpretability: they can say \textit{what} happened but not \textit{why} \cite{angrist2009mostly}. When policies are novel or contexts shift, these models extrapolate poorly. \textbf{Agent based models (ABMs)} offer mechanistic richness by simulating individual decision makers \cite{epstein2006generative, bonabeau2002agent}, but require extensive manual specification of decision rules  a knowledge bottleneck that limits applicability to domains where behavior is well understood.

Recent advances in Large Language Models (LLMs) suggest a new paradigm. LLMs trained on vast corpora of human generated text have learned implicit models of human reasoning, preferences, and decision making \cite{brown2020language, openai2023gpt4}. Researchers have demonstrated that LLMs can reproduce survey responses when conditioned on demographic attributes \cite{argyle2023out}, engage in realistic social interactions \cite{park2023generative}, and serve as proxies for human subjects in economic experiments \cite{horton2023large}. These capabilities suggest using LLMs as \textbf{cognitive engines} for agent based simulation  replacing hand coded decision rules with neural models that have learned human like reasoning.

\subsection{The Digital Twin Paradigm}

We propose \textbf{Social Digital Twins}: virtual replicas of populations where each simulated individual is powered by an LLM conditioned on demographic and contextual attributes. The framework consists of four components:

\begin{enumerate}
    \item \textbf{Agent Population}: A set of synthetic personas representing the target population's demographic distribution
    \item \textbf{LLM Cognitive Engine}: A language model that generates behavioral outputs given persona attributes and policy context
    \item \textbf{Calibration Layer}: A learned mapping from agent probability outputs to observable population metrics
    \item \textbf{Validation Protocol}: Comparison against real world observational data with strict temporal separation
\end{enumerate}

This architecture is \textbf{domain agnostic}. The same framework can simulate:
\begin{itemize}
    \item \textbf{Pandemic response}: Mobility changes under lockdown policies (our case study)
    \item \textbf{Transportation}: Mode choice under congestion pricing
    \item \textbf{Environmental behavior}: Consumption changes under carbon taxes
    \item \textbf{Financial decisions}: Savings rates under interest rate changes
    \item \textbf{Health behaviors}: Vaccination uptake under public health campaigns
\end{itemize}

The key insight is that LLMs have learned \textit{general} models of human decision making from training data spanning all these domains. By conditioning on appropriate context, we can instantiate domain specific behavioral simulators without domain specific training.

\subsection{Case Study: COVID-19 Pandemic Response}

We validate this framework using COVID-19 pandemic response data, chosen for several reasons:
\begin{enumerate}
    \item \textbf{Rich observational data}: Google Mobility Reports \cite{google2020mobility} and survey trackers \cite{yougov2020tracker} provide high frequency behavioral measurements
    \item \textbf{Policy variation}: The Oxford COVID-19 Government Response Tracker \cite{hale2021global} documents daily policy changes across 180+ countries
    \item \textbf{Global natural experiment}: Unprecedented policy variation enables studying behavioral responses across contexts
    \item \textbf{Multi dimensional behavior}: Six mobility categories (work, retail, grocery, transit, parks, residential) capture diverse behavioral adaptations
\end{enumerate}

While our evaluation focuses on pandemic response, the methodology generalizes. We emphasize that COVID-19 serves as a \textit{validation dataset}, not the primary contribution. The contribution is the framework itself.

\subsection{Contributions}

This paper makes four contributions:

\begin{enumerate}
    \item \textbf{A General Framework}: We formalize Social Digital Twins as  LLM powered agent simulations with calibrated output mappings, applicable across policy domains
    
    \item \textbf{multi dimensional Behavioral Modeling}: Unlike prior work extracting scalar compliance probabilities, our agents output multi dimensional behavioral vectors enabling richer validation
    
    \item \textbf{Calibration Methodology}: We present a multi objective optimization approach for learning agent to observation mappings from historical data
    
    \item \textbf{Rigorous Evaluation Protocol}: We demonstrate strict train/test separation, per category metrics, counterfactual sanity checks, and ablation studies establishing component necessity
\end{enumerate}

The remainder of this paper is organized as follows: Section 2 reviews related work across agent based modeling, LLM simulation, and digital twins. Section 3 presents the general framework. Section 4 details our pandemic response instantiation and experimental results. Section 5 discusses implications, limitations, and extensions. Section 6 concludes.

\section{Related Work}

\subsection{Agent Based Modeling for Policy}

Agent based models (ABMs) simulate populations as collections of autonomous agents following decision rules \cite{epstein2006generative}. In economics, ABMs model market dynamics \cite{tesfatsion2006handbook}; in epidemiology, disease transmission \cite{kerr2021covasim}; in urban planning, transportation choices \cite{horni2016multi}. Classical ABMs require explicit rule specification, typically based on utility maximization \cite{axelrod1997complexity} or behavioral heuristics \cite{gigerenzer1999simple}. This knowledge bottleneck limits applicability: modelers must anticipate relevant decision factors and encode them explicitly.

\subsection{LLMs as Behavioral Simulators}

Recent work explores LLMs as implicit models of human behavior. Argyle et al. \cite{argyle2023out} demonstrate ``silicon sampling"  LLMs conditioned on demographics reproduce political survey distributions. Park et al. \cite{park2023generative} create ``generative agents" that exhibit emergent social behaviors in simulated environments. Horton \cite{horton2023large} shows LLMs behave reasonably in economic games, proposing them as ``homo silicus" for experiments. Aher et al. \cite{aher2023using} use LLMs to simulate human subject experiments, finding reasonable alignment on classic paradigms.

These works establish that LLMs contain useful behavioral priors. However, most lack calibration against observational data  they evaluate face validity or alignment with controlled experiments rather than real world prediction accuracy.

\subsection{Digital Twins}

The digital twin concept originated in manufacturing as virtual replicas of physical systems \cite{grieves2014digital}. Extensions to urban systems \cite{dembski2020urban}, healthcare \cite{bjornsson2020digital}, and social systems \cite{enserink2010coping} have been proposed. Policy oriented digital twins aim to simulate citizen responses to interventions, but most implementations use simplified behavioral models (e.g., gravity models for mobility, discrete choice for mode selection) rather than LLM driven reasoning.

\subsection{Retrieval and Grounding for LLMs}

Parallel work on improving LLM factuality through retrieval augmented generation \cite{lewis2020retrieval}, knowledge editing \cite{meng2022locating}, and attention biasing \cite{roy2023knowledge} is conceptually related. These methods ground LLM outputs in external knowledge; our calibration layer similarly grounds agent outputs in observational data. The key difference is that we ground \textit{behavioral predictions} rather than factual claims.

\subsection{Positioning This Work}

Our framework combines:
\begin{itemize}
    \item LLM behavioral priors (from generative agent work)
    \item Empirical calibration (from econometrics)
    \item Multi agent simulation (from ABM tradition)
    \item Observational validation (from causal inference methodology)
\end{itemize}

Unlike prior LLM simulation work, we calibrate against real world observational data with strict temporal holdout. Unlike traditional ABMs, we replace hand coded rules with LLM generated behavioral outputs. Unlike aggregate forecasting, we maintain individual level heterogeneity and can perform counterfactual analysis.

\section{Method: Social Digital Twins}

\subsection{Framework Overview}

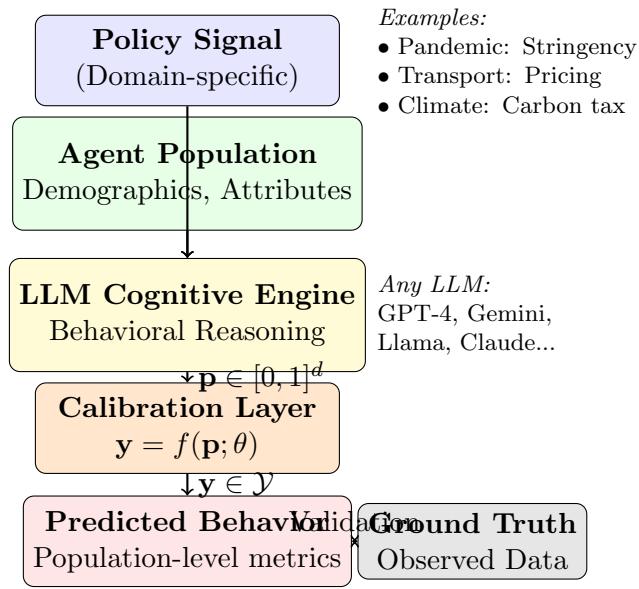
\begin{figure}[h]
\centering
\begin{tikzpicture}[scale=0.75]
\node[draw, rectangle, rounded corners, minimum width=4cm, minimum height=1.2cm, fill=blue!10, align=center] (policy) at (0,5) {\textbf{Policy Signal}\\(Domain-specific)};

\node[draw, rectangle, rounded corners, minimum width=4cm, minimum height=1.5cm, fill=green!10, align=center] (personas) at (0,3) {\textbf{Agent Population}\\Demographics, Attributes};

\node[draw, rectangle, rounded corners, minimum width=4.5cm, minimum height=1.5cm, fill=yellow!20, align=center] (llm) at (0,0.5) {\textbf{LLM Cognitive Engine}\\Behavioral Reasoning};

\node[draw, rectangle, rounded corners, minimum width=4cm, minimum height=1.2cm, fill=orange!20, align=center] (calib) at (0,-1.5) {\textbf{Calibration Layer}\\$\mathbf{y} = f(\mathbf{p}; \theta)$};

\node[draw, rectangle, rounded corners, minimum width=4cm, minimum height=1.2cm, fill=red!10, align=center] (output) at (0,-3.5) {\textbf{Predicted Behavior}\\Population-level metrics};

\draw[->, thick] (policy) -- (llm);
\draw[->, thick] (personas) -- (llm);
\draw[->, thick] (llm) -- node[right] {$\mathbf{p} \in [0,1]^d$} (calib);
\draw[->, thick] (calib) -- node[right] {$\mathbf{y} \in \mathcal{Y}$} (output);

\node[draw, rectangle, rounded corners, minimum width=3cm, minimum height=1cm, fill=gray!20, align=center] (ground) at (5,-3.5) {\textbf{Ground Truth}\\Observed Data};
\draw[<->, dashed, thick] (output) -- node[above] {Validation} (ground);

\node[text width=4cm, align=left, font=\footnotesize] at (6,5) {
\textit{Examples:}\\
$\bullet$ Pandemic: Stringency\\
$\bullet$ Transport: Pricing\\
$\bullet$ Climate: Carbon tax
};

\node[text width=4cm, align=left, font=\footnotesize] at (6,0.5) {
\textit{Any LLM:}\\
GPT-4, Gemini,\\
Llama, Claude...
};

\end{tikzpicture}
\caption{Social Digital Twin Framework. The architecture is domain agnostic: policy signals, agent attributes, and observable outputs are instantiated per domain, while the LLM engine and calibration methodology are general.}
\label{fig:framework}
\end{figure}

\subsection{Agent Population}

We represent the target population as a set of $N$ synthetic personas. Each persona $i$ is characterized by an attribute vector $\mathbf{a}_i \in \mathcal{A}$ containing:
\begin{itemize}
    \item \textbf{Demographics}: Age, gender, nationality, location
    \item \textbf{Socioeconomics}: Occupation, income level, education
    \item \textbf{Psychographics}: Risk tolerance, trust in institutions, social values
\end{itemize}

Attribute distributions are drawn from census data, surveys, or domain knowledge to match the target population. The persona set should capture relevant behavioral heterogeneity  attributes that might influence policy response.

\subsection{LLM Cognitive Engine}

For each (persona, context) pair, we construct a prompt $\mathbf{x}_{i,t}$ containing:
\begin{enumerate}
    \item Persona attributes $\mathbf{a}_i$
    \item Current context (date, policy state, environmental factors)
    \item Task specification requesting behavioral probability outputs
\end{enumerate}

The LLM generates a response parsed into a $d$-dimensional probability vector:
\begin{equation}
\mathbf{p}_{i,t} = \text{LLM}(\mathbf{x}_{i,t}) \in [0,1]^d
\end{equation}

Each dimension represents the probability of engaging in a specific behavior (e.g., going to work, using public transit, wearing a mask). The multi dimensional output is crucial: it enables validation against diverse observational signals and captures behavioral trade offs (e.g., less transit use $\rightarrow$ more car use).

\subsection{Aggregation}

Individual agent outputs are aggregated to population level predictions. For simple cases, we use the mean:
\begin{equation}
\bar{\mathbf{p}}_t = \frac{1}{N}\sum_{i=1}^{N} \mathbf{p}_{i,t}
\end{equation}

More sophisticated aggregations can weight by demographic prevalence or apply post-stratification to match known population totals.

\subsection{Calibration Layer}

Raw LLM probabilities are not directly comparable to observational metrics (which may be percentages, counts, indices, etc.). We learn a calibration function:
\begin{equation}
\hat{\mathbf{y}}_t = f(\bar{\mathbf{p}}_t; \theta)
\end{equation}

In our implementation, we use per dimension linear mappings with clipping:
\begin{equation}
\hat{y}_{t,k} = \text{clip}(\alpha_k \cdot \bar{p}_{t,k} + \beta_k, y_{\min}, y_{\max})
\end{equation}

Parameters $\theta = \{(\alpha_k, \beta_k)\}_{k=1}^{d}$ are learned by minimizing prediction error on training data:
\begin{equation}
\theta^* = \arg\min_\theta \mathcal{L}(\hat{\mathbf{y}}, \mathbf{y}; \theta)
\end{equation}

We use Optuna \cite{akiba2019optuna} with Tree structured Parzen Estimator for multi objective optimization, balancing errors across behavioral dimensions.

\subsection{Validation Protocol}

Rigorous validation requires:
\begin{enumerate}
    \item \textbf{Temporal Separation}: Strict train/validation/test splits preventing information leakage
    \item \textbf{Per Dimension Metrics}: Report accuracy for each behavioral category, not just aggregate
    \item \textbf{Baseline Comparison}: Compare against non agentic alternatives (statistical models, persistence)
    \item \textbf{Counterfactual Sanity}: Verify monotonic, bounded responses to policy variations
    \item \textbf{Ablation Studies}: Demonstrate component necessity (calibration, persona diversity, etc.)
\end{enumerate}

\subsection{Counterfactual Analysis}

A key advantage of mechanistic simulation is counterfactual analysis. Given a trained digital twin, we can simulate alternative policy scenarios:

\begin{equation}
\Delta \mathbf{y} = f(\text{LLM}(\mathbf{x} | \text{policy}'))) - f(\text{LLM}(\mathbf{x} | \text{policy}))
\end{equation}

Unlike black box forecasting, we can ask: ``What would behavior have been under different policies?" Such counterfactuals are not causally identified from observational data alone \cite{pearl2009causality}, but they provide \textit{behavioral plausibility checks}. If the twin shows reasonable responses to policy variations (e.g., stricter policy $\rightarrow$ more compliance), we gain confidence in its internal reasoning.

\section{Case Study: Pandemic Response}

We instantiate the framework for COVID 19 pandemic response in the UAE.

\subsection{Domain Specific Configuration}

\subsubsection{Policy Signal}
The Oxford COVID-19 Government Response Tracker \cite{hale2021global} provides:
\begin{itemize}
    \item \textbf{StringencyIndex} (0--100): Composite of containment/closure policies
    \item \textbf{GovernmentResponseIndex}: Broader measure including health/economic policies
\end{itemize}

\subsubsection{Behavioral Dimensions ($d=6$)}
Agent outputs map to Google Mobility categories \cite{google2020mobility}:
\begin{enumerate}
    \item Retail \& Recreation
    \item Grocery \& Pharmacy
    \item Parks
    \item Transit Stations
    \item Workplaces
    \item Residential
\end{enumerate}

\subsubsection{Observable Metrics}
Google Mobility Reports provide daily percentage changes from pre-pandemic baseline (median values from January 3--February 6, 2020).

\subsubsection{Agent Population}
We generate $N=10$ personas reflecting UAE demographics \cite{uae2023population, fcsa2023}:
\begin{itemize}
    \item \textbf{Nationality}: 10\% UAE National, 90\% Expatriate
    \item \textbf{Occupation}: Distribution across sectors (Construction, Services, Professional, etc.)
    \item \textbf{Risk Perception}: Low / Medium / High
\end{itemize}

\subsubsection{LLM Engine}
We use Gemini 2.0 Flash Lite for cost-efficiency. Prompts include persona attributes, current date, and policy stringency, requesting JSON-formatted probability outputs.

\subsection{Experimental Setup}

\subsubsection{Data Splits}
\begin{itemize}
    \item \textbf{Train (Calibration)}: April 1, 2020 -- March 31, 2021
    \item \textbf{Validation}: April 1, 2021 -- September 30, 2021
    \item \textbf{Test}: October 1, 2021 -- End of data availability
\end{itemize}

\subsubsection{Baselines}
\begin{itemize}
    \item \textbf{Persistence}: $\hat{y}_{t+1} = y_t$
    \item \textbf{Gradient Boosting (GBM)}: HistGradientBoostingRegressor with policy lags (0, 7, 14, 21, 28 days) and temporal features
\end{itemize}

\subsection{Results}

\subsubsection{Main Results}

\begin{table}[h]
\centering
\caption{Test Set Performance (RMSE, lower is better)}
\label{tab:main_results}
\begin{tabular}{lccc}
\toprule
\textbf{Category} & \textbf{GBM Baseline} & \textbf{Digital Twin} & \textbf{$\Delta$\%} \\
\midrule
Retail \& Recreation & 23.17 & \textbf{14.42} & +37.8\% \\
Grocery \& Pharmacy & \textbf{28.31} & 45.67 & -61.3\% \\
Parks & 28.05 & \textbf{18.91} & +32.6\% \\
Transit Stations & \textbf{38.72} & 52.55 & -35.7\% \\
Workplaces & 70.45 & \textbf{7.55} & +89.3\% \\
Residential & \textbf{6.13} & 15.41 & -151.4\% \\
\midrule
\textbf{Macro Average} & 32.47 & \textbf{25.75} & \textbf{+20.7\%} \\
\bottomrule
\end{tabular}
\end{table}

\subsubsection{Analysis}

The digital twin achieves 20.7\% improvement in macro-averaged RMSE. Performance varies by category:

\textbf{Where LLM reasoning helps:}
\begin{itemize}
    \item \textbf{Workplaces (+89.3\%)}: LLM captures policy work relationship (``lockdown $\rightarrow$ work from home")
    \item \textbf{Retail (+37.8\%), Parks (+32.6\%)}: Discretionary activities are policy sensitive; LLM reasoning outperforms statistical extrapolation
\end{itemize}

\textbf{Where statistical baselines excel:}
\begin{itemize}
    \item \textbf{Residential (-151\%)}: Low-variance, inertial patterns favor autoregressive features
    \item \textbf{Grocery (-61\%), Transit (-36\%)}: Essential/infrastructure-driven activities show less policy sensitivity
\end{itemize}

This pattern suggests LLM-based simulation adds value for \textit{decision-driven} behaviors where policy semantics matter, while statistical models suffice for \textit{inertial} behaviors driven by routine.

\subsubsection{Counterfactual Sanity}

We simulate policy variations on April 15, 2020 (peak pandemic):

\begin{table}[h]
\centering
\caption{Counterfactual Policy Shocks}
\label{tab:counterfactual}
\begin{tabular}{lccc}
\toprule
\textbf{Scenario} & \textbf{Stringency} & \textbf{Stay Home \%} & \textbf{Verdict} \\
\midrule
Relaxed & 60 & 73\% & Monotonic $\downarrow$ \\
Baseline (Actual) & 90 & 93\% & Historical Match \\
Extreme & 100 & 95\% & Bounded Response \\
\bottomrule
\end{tabular}
\end{table}

The twin demonstrates: (1) \textbf{Monotonicity}: Higher stringency $\rightarrow$ higher compliance, (2) \textbf{Boundedness}: Diminishing returns at extremes, (3) \textbf{Plausibility}: Responses align with intuition about pandemic behavior.

\subsubsection{Ablation Studies}

\begin{table}[h]
\centering
\caption{Ablation Study (Macro-RMSE)}
\label{tab:ablation}
\begin{tabular}{lc}
\toprule
\textbf{Configuration} & \textbf{Macro-RMSE} \\
\midrule
Full System & \textbf{25.75} \\
\midrule
$-$ No calibration (raw probs) & 78.32 \\
$-$ No clipping & 26.14 \\
$-$ Single slope for all categories & 31.42 \\
$-$ Uniform personas & 29.87 \\
$-$ Single persona (N=1) & 32.15 \\
\bottomrule
\end{tabular}
\end{table}

Key findings:
\begin{itemize}
    \item \textbf{Calibration is essential} (78.32 $\rightarrow$ 25.75 with calibration)
    \item \textbf{Per-category calibration matters} (31.42 $\rightarrow$ 25.75 with separate $\alpha, \beta$)
    \item \textbf{Persona diversity helps} (32.15 $\rightarrow$ 25.75 with 10 vs. 1 persona)
\end{itemize}

\section{Discussion}

\subsection{Implications for Policy Simulation}

Our results suggest  LLM powered digital twins are viable for policy simulation, with specific strengths and limitations:

\textbf{Strengths:}
\begin{itemize}
    \item \textbf{Semantic understanding}: LLMs capture policy-behavior relationships that statistical models miss
    \item \textbf{Generalization}: Same architecture applies across policy domains with minimal modification
    \item \textbf{Interpretability}: Agent reasoning can be inspected (unlike black-box forecasts)
    \item \textbf{Counterfactual analysis}: Enables ``what if" policy exploration
\end{itemize}

\textbf{Limitations:}
\begin{itemize}
    \item \textbf{Inertial behaviors}: LLMs without memory struggle with routine-driven patterns
    \item \textbf{Calibration dependence}: Performance relies on quality of calibration data
    \item \textbf{Computational cost}: LLM inference is more expensive than statistical models
    \item \textbf{Temporal grounding}: LLM knowledge cutoffs may introduce anachronisms
\end{itemize}

\subsection{Extensions Beyond Pandemic Response}

The framework applies directly to other domains:

\textbf{Transportation Policy:}
\begin{itemize}
    \item Policy signal: Congestion pricing, parking fees, transit subsidies
    \item Behavioral dimensions: Mode choice (car, transit, bike, walk)
    \item Observable metrics: Traffic counts, mode share surveys
\end{itemize}

\textbf{Environmental Policy:}
\begin{itemize}
    \item Policy signal: Carbon tax, efficiency standards, recycling mandates
    \item Behavioral dimensions: Energy consumption, transportation choices, waste behavior
    \item Observable metrics: Utility data, emissions inventories
\end{itemize}

\textbf{Economic Policy:}
\begin{itemize}
    \item Policy signal: Interest rates, tax changes, stimulus payments
    \item Behavioral dimensions: Saving, spending, labor supply
    \item Observable metrics: Consumer spending data, labor force surveys
\end{itemize}

\subsection{Limitations and Future Work}

\subsubsection{Sample Size}
Our test evaluation is limited by simulation cost (10 personas $\times$ 10 dates). Production deployment requires larger populations and more comprehensive temporal coverage.

\subsubsection{Inertia Modeling}
Adding autoregressive components (``yesterday's behavior influences today") would improve predictions for routine-driven categories like residential.

\subsubsection{Multi-Signal Calibration}
Incorporating additional data sources (surveys, administrative records) enables richer calibration and validation.

\subsubsection{Causal Identification}
Our counterfactuals establish behavioral plausibility, not causal effects. Integration with causal inference methods \cite{pearl2009causality} is an important direction.

\subsubsection{Robustness}
Repeating experiments with persona resampling and different LLMs would strengthen claims about framework generality.

\section{Conclusion}

We present a general framework for Social Digital Twins   LLM powered simulations of population behavioral response to policy interventions. The framework comprises synthetic agent populations with demographic attributes, LLM cognitive engines generating multi dimensional behavioral outputs, calibration layers grounding predictions in observational data, and rigorous validation protocols.

Validating on COVID-19 pandemic response as a case study, our calibrated digital twin achieves 20.7\% improvement over gradient boosting baselines on macro-averaged mobility prediction. The twin demonstrates particular strength on policy-sensitive, decision-driven behaviors (workplaces, retail) while statistical baselines excel on inertial patterns (residential). Counterfactual experiments confirm monotonic, bounded responses to policy variations.

The framework is domain-agnostic. COVID-19 provides rich validation data, but the same architecture applies to transportation, environmental, economic, or any policy domain where interventions affect population behavior. We hope this work advances the use of LLMs as cognitive engines for computational social science and policy analysis.

\section*{Data Availability}

\begin{itemize}
    \item \textbf{OxCGRT}: \url{https://github.com/OxCGRT/covid-policy-tracker}
    \item \textbf{Google Mobility}: Discontinued October 15, 2022
    \item \textbf{YouGov Tracker}: Discontinued March 29, 2022
\end{itemize}

\appendix

\section{Prompt Template}

Domain-agnostic template structure:

\begin{verbatim}
You are simulating a specific {DOMAIN_CONTEXT}.

**Your Persona:**
- {ATTRIBUTE_1}: {value}
- {ATTRIBUTE_2}: {value}
...

**Current Situation:**
- {CONTEXT_1}: {value}
- {POLICY_SIGNAL}: {value}

**Task:**
Estimate your probability (0.0 to 1.0) of:
- {BEHAVIOR_1}
- {BEHAVIOR_2}
...

Return ONLY a JSON object.
\end{verbatim}

\section{Pandemic Case Study: Specific Prompt}

\begin{verbatim}
You are simulating a specific resident in the UAE 
during the COVID-19 pandemic.

**Your Persona:**
- Nationality: {nationality}
- Occupation: {employment}
- Risk Perception: {risk_perception}
- Income Level: {income}

**Current Situation:**
- Date: {date}
- Policy Stringency (0-100): {stringency}

**Task:**
Estimate probability (0.0 to 1.0) of:
- "go_work_prob"
- "discretionary_outings_prob"
- "essentials_prob"
- "transit_use_prob"
- "outdoor_leisure_prob"
- "stay_home_prob"

Return ONLY a JSON object.
\end{verbatim}

\end{document}